\documentclass[letterpaper]{article} 
\usepackage{aaai23}  
\usepackage{times}  
\usepackage{helvet}  
\usepackage{courier}  
\usepackage[hyphens]{url}  
\usepackage{graphicx} 
\usepackage{natbib}  
\usepackage{caption} 
\usepackage{algorithm}
\usepackage{algorithmic}
\usepackage{amsfonts} 
\usepackage{amsmath}
\usepackage{amssymb}
\usepackage{dsfont}
\usepackage{xcolor}
\usepackage{multirow}
\usepackage{multicol}
\usepackage{booktabs}
\usepackage{makecell}
\usepackage{arydshln}
\usepackage{cuted}
\usepackage{newfloat}
\usepackage{listings}
\DeclareCaptionStyle{ruled}{labelfont=normalfont,labelsep=colon,strut=off} 
\floatstyle{ruled}
\newfloat{listing}{tb}{lst}{}
\floatname{listing}{Listing}
\title{Joint Self-Supervised Image-Volume Representation Learning \\ with 
Intra-Inter Contrastive Clustering}
\author{
    Duy M. H. Nguyen\textsuperscript{\rm 1,10*},
    Hoang Nguyen\textsuperscript{\rm 2},
    Mai T. N. Truong\textsuperscript{\rm 3},
    Tri Cao\textsuperscript{\rm 2},
    Binh T. Nguyen\textsuperscript{\rm 2} \\
    Nhat Ho\textsuperscript{\rm 4},
    Paul Swoboda\textsuperscript{\rm 5},
    Shadi Albarqouni\,\textsuperscript{\rm 6,7},
    Pengtao Xie\textsuperscript{\rm 8},
    Daniel Sonntag\textsuperscript{\rm 9,10}
}
\affiliations{
    \textsuperscript{\rm 1}University of Stuttgart,
    \textsuperscript{\rm 2}AISIA Lab, University of Science -VNU HCM, 
    \textsuperscript{\rm 3}Dongguk University \\
    \textsuperscript{\rm 4}University of Texas at Austin 
    \textsuperscript{\rm 5}Max Planck Institute for Informatics,
    \textsuperscript{\rm 6}Helmholtz AI, Helmholtz Munich,
    \textsuperscript{\rm 7}University of Bonn,
    \textsuperscript{\rm 8}University of California San Diego,
    \textsuperscript{\rm 9}Oldenburg University,
    \textsuperscript{\rm 10}German Research Center for Artificial Intelligence \\ 
    
    \textsuperscript{\rm *}Corresponding email: ho\_minh\_duy.nguyen@dfki.de
}

\begin{document}

\maketitle

\begin{abstract}
Collecting large-scale medical datasets with fully annotated samples for training of deep networks is prohibitively expensive, especially for 3D volume data. Recent breakthroughs in self-supervised learning (SSL) offer the ability to overcome the lack of labeled training samples by learning feature representations from unlabeled data. However, most current SSL techniques in the medical field have been designed for either 2D images or 3D volumes. In practice, this restricts the capability to fully leverage unlabeled data from numerous sources, which may include both 2D and 3D data. Additionally, the use of these pre-trained networks is constrained to downstream tasks with compatible data dimensions.
In this paper, we propose a novel framework for unsupervised joint learning on 2D and 3D data modalities. Given a set of 2D images or 2D slices extracted from 3D volumes, we construct an SSL task based on a 2D contrastive clustering problem for distinct classes. The 3D volumes are exploited by computing vectored embedding at each slice and then assembling a holistic feature through deformable self-attention mechanisms in Transformer, allowing incorporating long-range dependencies between slices inside 3D volumes. These holistic features are further utilized to define a novel 3D clustering agreement-based SSL task and masking embedding prediction inspired by pre-trained language models. Experiments on downstream tasks, such as 3D brain segmentation, lung nodule detection, 3D heart structures segmentation, and abnormal chest X-ray detection, demonstrate the effectiveness of our joint 2D and 3D SSL approach. We improve plain 2D Deep-ClusterV2 and SwAV by a significant margin and also surpass various modern 2D and 3D SSL approaches.
\end{abstract}

\section{Introduction}
Creating large-scale medical image datasets for training neural networks is a major obstacle due to the complexity of data acquisition, expensive annotations, and privacy concerns \citep{cheplygina2019not,kaissis2020secure}.
To alleviate these challenges, a conventional approach is to train deep networks, e.g., ResNet-50 \cite{he2016deep}, on large-scale natural image datasets such as ImageNet~\cite{deng2009imagenet} and subsequently fine-tune them on the target medical domain. However, such schemes are sub-optimal due to the large domain discrepancy between natural images and medical data~\cite{raghu2019transfusion,nguyen2022tatl}.
This has motivated other techniques for collecting annotated medical datasets across domains and training networks using full \cite{gibson2018niftynet,chen2019med3d} or semi-supervision \cite{wang2020focalmix}.
Nevertheless, the amount of acquired relevant training data in this manner is still limited, which significantly limits the performance of deep neural networks.

\begin{figure}
    \includegraphics[width=0.45\textwidth]{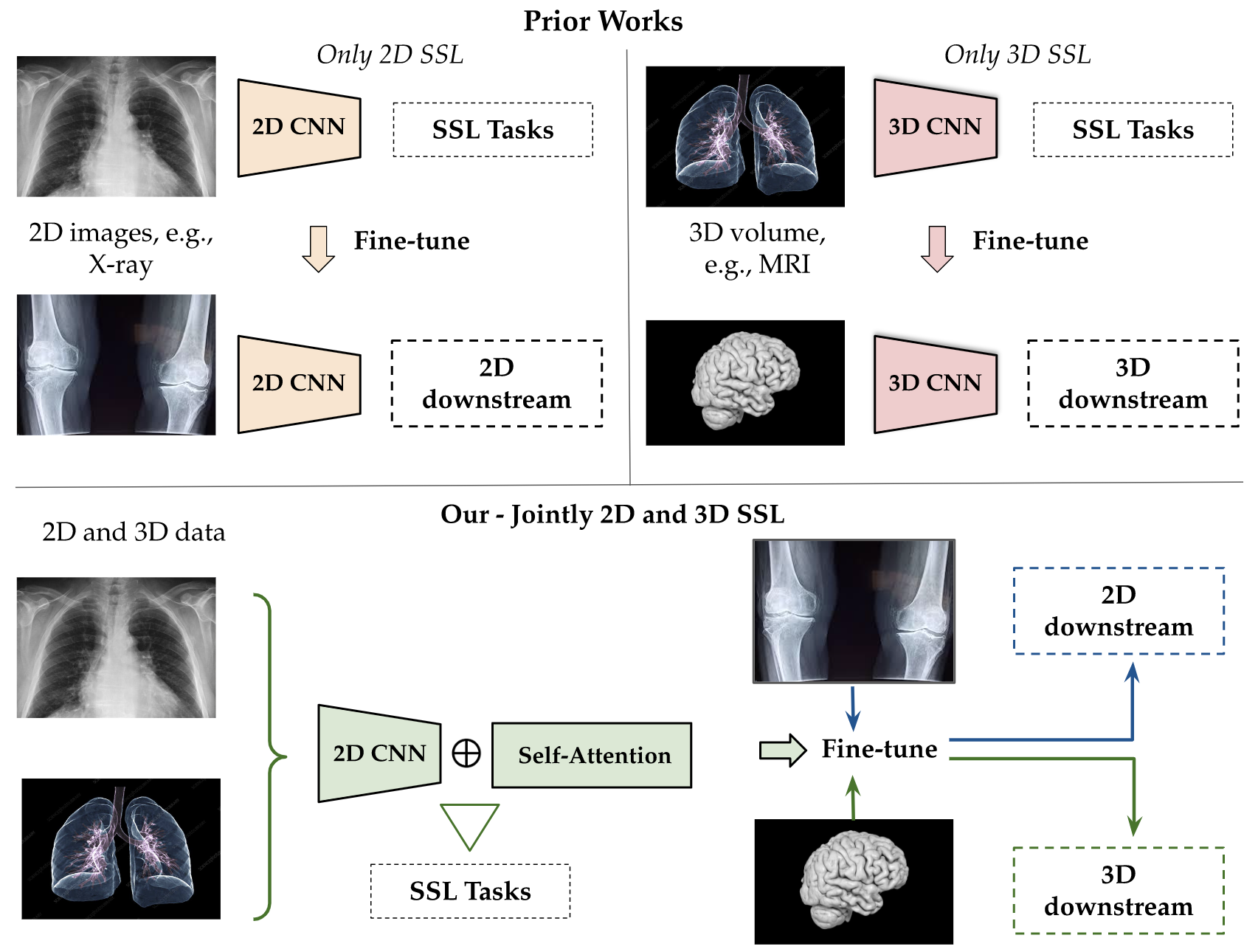}
    \vspace{-0.1in}
    \caption{\small{The main distinctions between our work and prior studies on 2D and 3D self-supervised learning. We can learn representations from diverse data and the pre-trained weights can be transferred for both 2D and 3D downstream tasks.}}
    \vspace{-0.2in}
    \label{fig:teaser}
\end{figure}

Self-supervised learning (SSL) has recently emerged as a new trend in medical imaging due to its ability in obtaining feature representations from unlabeled data by solving proxy tasks, which can be broadly categorized into \textit{generative}~\cite{chen2019self} and \textit{discriminative} ones~\cite{chen2020simple,he2020momentum}.
Discriminative SSL can be further separated into three directions: instance level-based methods  \cite{zbontar2021barlow,caron2021emerging}, contrastive learning-based methods \cite{he2020momentum,chen2021empirical} and clustering-based methods \cite{caron2020unsupervised,li2021prototypical}. Depending on a specific 2D, e.g., X-ray images or 3D magnetic resonance imaging (MRI) application, variations of these methods can be modified using 3D convolutional neural networks (CNNs) or Transformer architectures \cite{taleb20203d,haghighi2021transferable,tang2022self}.

However, all aforementioned SSL methods have been designed to learn on either 2D or 3D data modalities. As a result, they suffer from two major limitations. First, the ability to exploit unlabeled data from multiple source domains, which commonly occurs in medical data, is restricted. For instance, 3D CNN-based SSL methods can not use X-rays, digital retinal, and dermoscopic images taken from lung, eye retina, and skin lesions, respectively. Although 2D CNN-based SSL methods can process 3D volumes slice-by-slice along a specific plane (either sagittal, coronal, or horizontal) \cite{nguyen2022asmcnn,jun2021medical}, these approaches do not capture long-range inter-slice correlations and thus may result in inferior performance in 3D applications. Second, using a purely 2D or 3D strategy limits the fine-tuning phase since the pre-trained models are only applicable for downstream tasks with the same dimensionality. For instance, pre-trained 3D-CNN cannot handle object detection \cite{nguyen2021attention,nguyen2022vindr} (Table \ref{tab:unseen-domain-univer}, third column) and similarly pre-trained 2D-CNN might not be usable for 3D classification tasks (Table \ref{tab:see-domains}, second column). 

In this work, we propose a novel technique to overcome those barriers by presenting a hybrid SSL architecture harnessing both 2D and 3D medical data. The method has the following properties.
First, it is built on top of cutting-edge 2D SSL baselines while reserving designed CNN architecture, benefiting from the latest advancements of SSL in natural images.
Second, when applied to 3D data, we formulate both intra-dependencies inside slices and long-range inter-dependencies across slices, resulting in more complex contrastive cues that force the network to seek associated local and global feature representations.

Specifically, we compose a joint image-volume representation learning comprising a 2D CNN (ResNet-50) to extract feature embedding at the image level and a deformable attention transformer \cite{zhu2020deformable,liu2021swin,xia2022vision} to express correlations among local slices, aiming to derive  a holistic representation at the 3D volume level. Unlike standard attentions in Transformer \cite{vaswani2017attention,dosovitskiy2020image} which treat all attention positions equally, our deformable mechanism pays attention to only a flexible small set of major slices conditioned on input data. This largely reduces computational complexity and permits handling the multi-scale feature maps which are desired properties in medical downstream tasks. 

The proposed method is  trained on SSL tasks utilizing both current 2D SSL methodologies and our two novel 3D pre-text tasks. To this end, we employ two state-of-the-art contrastive clustering-based SSL approaches, Deep-Cluster-V2 ~\cite{caron2018deep} and SwAV~\cite{caron2020unsupervised}. With each baseline, we first perform the relevant 2D proxy tasks based on an \textit{agreement clustering for 2D slices} taken from 3D volumes. We next compute multi-level features at each slice within a 3D volume encoded with their positions and feed them into the deformable transformer. The global embedded features derived from this transformer are employed to define an \textit{agreement clustering for 3D volumes} and a \textit{masked encoding feature prediction} motivated by the success of the language model BERT~\cite{devlin2018bert}. By optimizing these conditions, intuitively we are able to learn feature extractors at the local- and global-level in a constraint manner, resulting in consistent cues and improved performance in downstream tasks. Furthermore, the pre-trained networks are adaptable with data dimensional compatibility by employing the 2D CNN for 2D tasks or the hybrid 2D CNN- Transformer architectures for 3D tasks.

In summary, we make the following contributions. First, we present an SSL framework capable of using various data dimensions and producing versatile pre-trained weights for both 2D and 3D downstream applications (Figure \ref{fig:teaser}). Second, we introduce the deformable self-attention mechanisms which utilize multi-level feature maps and capture flexible correlations between 2D slices, resulting in a powerful global feature representation. On top of this, we developed the
novel 3D agreement clustering extended from the earlier 2D clustering problem as well as proposed the masking embedding prediction. Finally, extensive experiments on public benchmarks confirmed that we improve state-of-the-art 2D baselines and surpass several latest SSL competitors based on CNN or Transformer.

\section{Related Work}
\paragraph{\textbf{Self-supervised Learning in Medical Image Analysis}}
Our work is closely related to instance-based constrative learning and unsupervised contrastive clustering. The \textit{instance-based contrastive methods} seek an embedding space where transformed samples, e.g., crops, drawn from the same instance, e.g., image, are pulled closer, and samples from distinct instances are pushed far away. The contrastive loss is constructed based on positive and negative feature pairs generated by various approaches, such as memory bank \cite{wu2018unsupervised}, end-to-end \cite{chen2020simple}, or momentum encoder \cite{chen2021empirical}. Despite achieving good performance in various settings, the instance-based method has crucial limitations in requiring a large negative batch size and choosing hard enough negative ones. The \textit{unsupervised contrastive clustering} \cite{caron2018deep,caron2020unsupervised} in other directions tries to learn representations based on groups of images with similar features rather than individual instances. For instance, SwAV \cite{caron2020unsupervised} simultaneously clusters the data while imposing consistency between cluster assignments generated for distinct augmentations of the same image. Currently, extensions on this direction have considered latent variables of centre points \cite{li2021prototypical}, multi-view clustering \cite{pan2021multi}, or mutual information \cite{do2021clustering}.  

In medical image analysis, several SSL methods have designed pre-text tasks based on 3D volume's properties such as reconstructing spatial context \cite{zhuang2019self}, random permutation prediction \cite{chen2019self}, self-discovery and self-restoration \cite{zhou2021models,haghighi2021transferable}. Some other efforts attempted to develop 3D CNN architecture while retaining defined SSL tasks on 2D CNN \cite{taleb20203d}. Another line of research considered the cross-domain training with two or more datasets, aiming to derive a generic invariant pre-trained model \cite{DBLP:journals/corr/abs-2010-06107}. Besides, existing methods also exploit the domain- and problem-specific cues such as structural similarity across 3D volumes in order to define global and local contrastive losses \cite{chaitanya2020contrastive,xie2020pgl}. However, most of these techniques have only been applied to 2D or 3D data, which are different from ours in terms of data usage and flexible pre-trained weights in downstream tasks (Figure \ref{fig:teaser}). 
\vspace{-0.05in}
\paragraph{\textbf{SSL Transformer in Medical Imaging}}
Vision transformers, adapted from sequence-to-sequence modeling in natural language processing, are initially used in image classification tasks \cite{dosovitskiy2020image}. In the context of SSL, 2D transformer-based methods such as Moco-v3 \cite{chen2021empirical} and DINO \cite{caron2021emerging} are also introduced and achieved promising performance. To elaborate 3D volumes, \citet{tang2022self} introduced a 3D transformer-based model comprising a Swin Transformer encoder \cite{liu2021swin} and skip connections. Likewise, \citet{xie2021unified} adapted a mixed 2D-3D Pyramid Vision Transformer architecture \cite{wang2021pyramid} to learn rich representations from diverse data. 

Compared with prior works in SSL \cite{caron2021emerging,tang2022self}, we employ Transformer to define the interaction between 2D slices inside a 3D volume rather than a fixed 2D or 3D network backbone, allowing us to adapt to varied data dimension downstream applications. Furthermore, we the first adapt deformable attention mechanism \cite{zhu2020deformable,liu2021swin,xia2022vision} for SSL, which currently are only validated performance in supervised learning. \citet{xie2021unified} shares the same ideas with us in jointly learning diverse unlabeled data; however, this method designs a specific SSL task while our 3D loss is extended directly from standard 2D cases. Also, we achieve similar or better performance compared with this baseline while using a smaller amount of unlabeled data.

\section{Methodology}
\begin{figure}[H]
    \centering
    \includegraphics[width=1.0\linewidth]{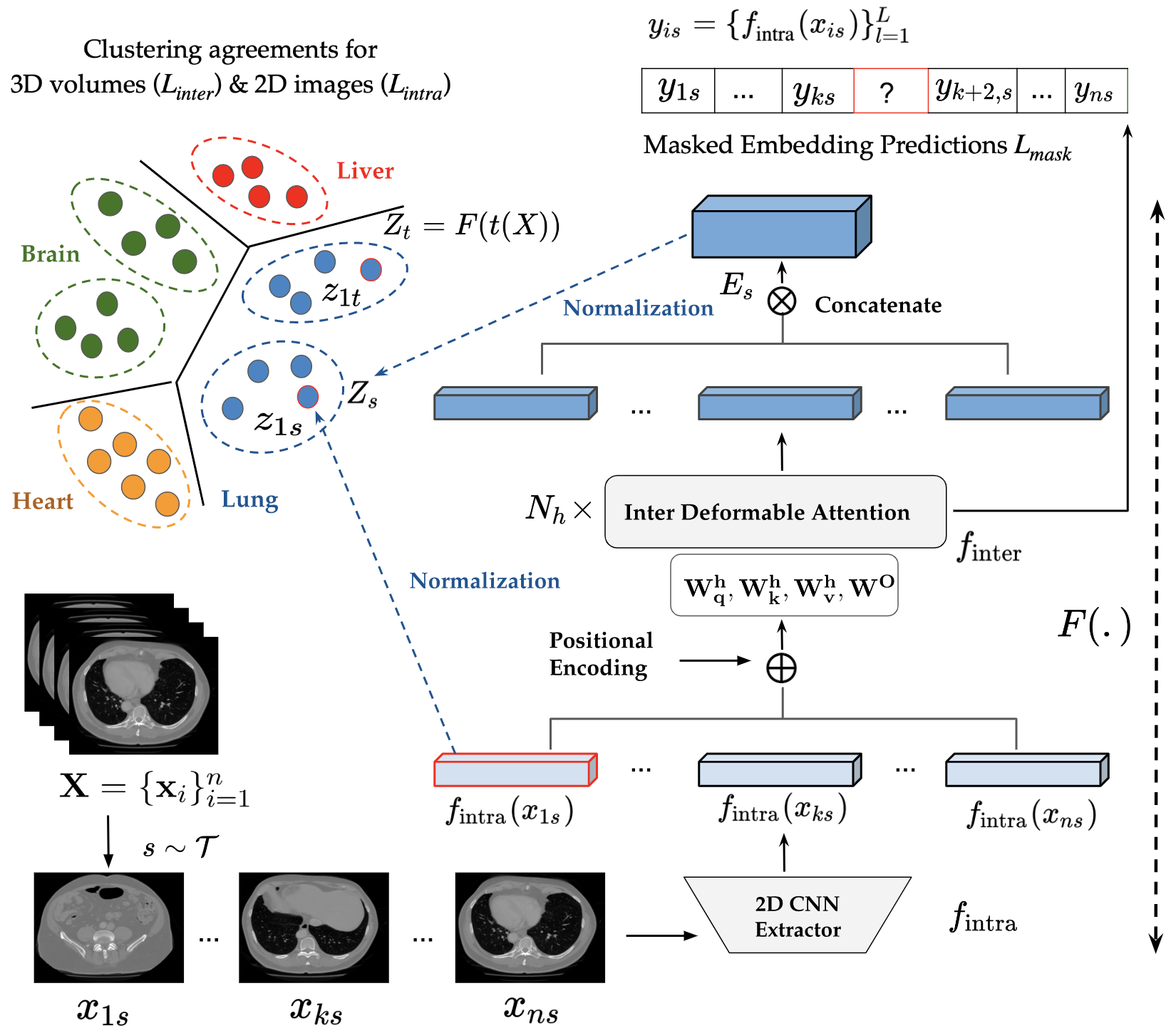}
    \vspace{-0.2in}
    \caption{
    \small{Overview of our joint SSL image-volume framework.
    Given a 3D volume $\mathbf{X}$ and a random transformation $s$, we compute the embedding feature for each slice using a 2D-CNN extractor $f_{\mathrm{intra}}$ and produce a global feature $Z_{s}$ through the Inter Deformable Attention $f_{\mathrm{inter}}$. Similarly, corresponding features can be derived from 2D and 3D augmented views of $\mathbf{X}$ by another transformation $t$. Through cluster agreement losses for 2D slices ($L_{\mathrm{intra}}$), e.g.\ between $z_{1s}$ and $z_{1t}$, and for 3D volumes between $Z_s$ and $Z_t$ ($L_{inter}$), feature representations can be learned. Additionally, we employ a masked feature embedding prediction given 2D slices' embedding outputs as an SSL task to capture data's long-term interdependence.}
    } 
    \label{fig:overview_method}
\end{figure}

Our approach is built on top of 2D contrastive clustering learning baselines including Deep-ClusterV2~\cite{caron2018deep} and SwAV~\cite{caron2020unsupervised}. Both approaches rely on clustering together features produced by neural network backbones. Deep-ClusterV2 forces each cluster to have roughly the same size. SwAV additionally imposes losses on assigning augmentations of an image into the same cluster. Below, we recapitulate the SwAV baseline and then show how it can be extended through the deformable self-attention~\cite{zhu2020deformable,xia2022vision} to 3D volumes.
Additionally, we introduce a new proxy task based on missing embedding prediction in order to make the designed architecture be stable under perturbations. An illustration of our approach can be seen in Figure~\ref{fig:overview_method}. A variation of our method using DeepCluster-V2 can be derived analogously.

\textbf{Notation:} We assume to be given $\mathrm{K}$ unlabeled datasets $\mathbb{D} = \{\mathrm{D_{1}}, \mathrm{D_{2}},..., \mathrm{D_{K}}\}$ consisting of instances $\mathrm{D_{i} = \{\mathbf{X_1},\,\mathbf{X_2},..., \mathbf{X_{m_i}}\}, i \in [1,\,K]}$, which include $\mathbf{m_i}$ 2D or 3D volumes $\mathbf{X_{j}},\, \mathbf{j} \in [\mathbf{1}, \mathbf{m_i}]$. Given a particular dataset $\mathrm{D} \in \mathbb{D}$, we assume that each 3D volume contains $n$ slices, i.e.\  $\forall \,\mathbf{X} \in \mathrm{D},\ \mathbf{X} =\{\mathbf{x}_i\}_{i=1}^{n}.$

\vspace{-0.2in}
\subsection{Clustering Agreement for 2D Images}
SwAV uses a proxy task for a ``swapped'' prediction problem in which the cluster assignment of a transformed image is to be found from the feature representation of another transformation of the same image and vice versa. In our framework, we refer to this proxy task as an \emph{intra-dependence} correlation since it learns only from 2D slices inside a 3D volume without taking into account correlations between different slices of the same volume. Below we formally specify the intra-dependence correlation.

Let $f_{\mathrm{intra}}$ be a CNN, e.g., ResNet-50 \cite{he2016deep}, extracting feature embeddings for each 2D slice $\mathbf{x}_i \in \mathbf{X}$. The cluster assignment matrix $\mathbf{C} = [c_1,\ldots,c_H]$ has columns $c_j$, each column corresponding to the feature representation of the $j$-th cluster, and $H$ is the number of hidden clusters. Given a 2D slice $\mathbf{x}_{i} \in \mathbf{X}$, we choose randomly two transformations $s,t \in T$, where $T$ is a set of pre-defined image transformations.
We apply $s$ and $t$ on $\mathbf{x}_{i}$ and obtain two augmented views $\mathbf{x}_{is},\, \mathbf{x}_{it}$.
Using $f_{\mathrm{intra}}$ and normalization gives us the respective features $\mathbf{z}_{it}$ and $\mathbf{z}_{is}$ (Figure \ref{fig:overview_method}), i.e.\ 
\begin{equation}
    \mathbf{z_{ik}} = f_{\mathrm{intra}}(\mathbf{x}_{ik})/||f_{\mathrm{intra}}(\mathbf{x}_{ik})||_2,\ k \in \{s,t\}.
    \label{eq:2d_augmented_view}
\end{equation}
These features are then used to find corresponding cluster assignments $\mathbf{q}_{it},\, \mathbf{q}_{is}$, i.e., the probability distribution over all clusters, called codes in SwAV. To find these codes, we sample a batch of size $B$ from slices of volumes coming from all datasets and optimize 
\begin{equation}
\scalebox{1.0}{$
\label{eq:matching}
    \max_{\mathbf{Q} \in \mathbb{Q}} \mathbf{Tr}(\mathbf{Q}^{T}\mathbf{C}^{T}\mathbf{Z}) + \epsilon H(\mathbf{Q}),$}
\end{equation}
where $\mathbf{Z} =[\mathbf{z}_1,...,\mathbf{z}_{2B}]$ is formed by adding features $z_{it},\,z_{is}$ of each $x_{i}$ in the batch $B$, the assignment matrix is $\mathbf{Q} = \left[\mathbf{q}_1,\ldots,\mathbf{q}_{2B} \right]$ and $\mathbb{Q} = \{ \mathbf{Q} \in \mathbb{R}_+^{K \times B} : \mathbf{Q} \mathds{1}_B = \frac{1}{K} \mathds{1}_K, \mathbf{Q} \mathds{1}_K = \frac{1}{B} \mathds{1}_B \}$ is the set of all possible assignment matrices such that slices  are assigned on average uniformly,
$H$ is the entropy function and $\epsilon$ is a hyper-parameter that controls the smoothness of the mapping. Since views coming from the same sample $\mathbf{x}_i$ should have features that are assigned to the same cluster, we formulate the intra-dependency code prediction loss
\begin{equation}
    L_{\mathrm{intra}}(\mathbf{z}_{it}, \mathbf{q}_{it}, \mathbf{z}_{is}, \mathbf{q}_{is}) = l(\mathbf{z}_{it}, \mathbf{q}_{is}) + l(\mathbf{z}_{is}, \mathbf{q}_{it})
\label{eq:loss_idea_intra}
\end{equation}
where the function $l(\mathbf{z}, \mathbf{q})$ quantifies the fit between feature $\mathbf{z}$ and code assignment $\mathbf{q}$ defined as
\begin{equation}
    l(\mathbf{z}_t, \mathbf{q}_s) = - \sum_{k} \mathbf{q}_{s}^{k}\log \mathbf{p}_{t}^{k} , \textrm{  where } \mathbf{p}_{t}^{k} = \frac{\exp(\frac{1}{\tau}\mathbf{z}_{t}^{T}\mathbf{c}_{k})}{\sum_{k'}\exp(\frac{1}{\tau}\mathbf{z}_{t}^{T}\mathbf{c}_{k'})}\,.
\label{eq:loss_one_side}
\end{equation}
Here $\tau$ is a hyper-parameter. 

Intuitively, if two features encode views coming from the same slice, the loss $l(\mathbf{z}_t, \mathbf{q}_s)$ in Eq.\,(\ref{eq:loss_one_side}) encourages their predicted clusters should be identical.
Finally, by optimizing Eq.\,(\ref{eq:loss_idea_intra}) over $\mathbf{x}_{i} \in \mathbf{X}$ we can learn feature representations $f_{\mathrm{intra}}$ and centroids $\mathbf{C}$ by minimizing
\begin{equation}
    L_{2D} =  \min_{f_{\mathrm{intra}}, \mathbf{C}} \mathrm{E}_{\mathbf{x}_{i} \in \mathbf{X}}\left[L_{\mathrm{intra}}(\mathbf{z}_{it}, \mathbf{q}_{it}, \mathbf{z}_{is}, \mathbf{q}_{is})\right],\ s,t\sim T.
    \label{eq:2D-clustering}
\end{equation}

\subsection{Clustering Agreement for 3D Volumes with Inter Deformable Attention}
\label{sub:self-attention}
In the presence of both unlabeled 2D and 3D data, we argue that the clustering agreement constraint in Eq.\eqref{eq:loss_one_side} should also hold for feature representations of \emph{different views of the 3D volume} (Figure \ref{fig:overview_method}). We call this agreement as an \textit{inter-dependence correlation}. It forces the feature representation to additionally consider long-range interactions among 2D slices inside a 3D volume (Eq.\eqref{eq:3D-feature}). To this end, we adapt the Transformer to aggregate local features computed by $f_{\mathrm{intra}}$ at each slice to form a holistic feature representation for a 3D volume. However the standard attention mechanisms in vanilla Transformer such as ViT \cite{dosovitskiy2020image} does not fit well in our setting when it permits excessive number of keys to contribute per query patch. As a result, the required memory and computational costs increase significantly as well as features can be influenced by irrelevant parts.

\begin{figure}[!htb]
\centering
\includegraphics[width=0.36\textwidth]{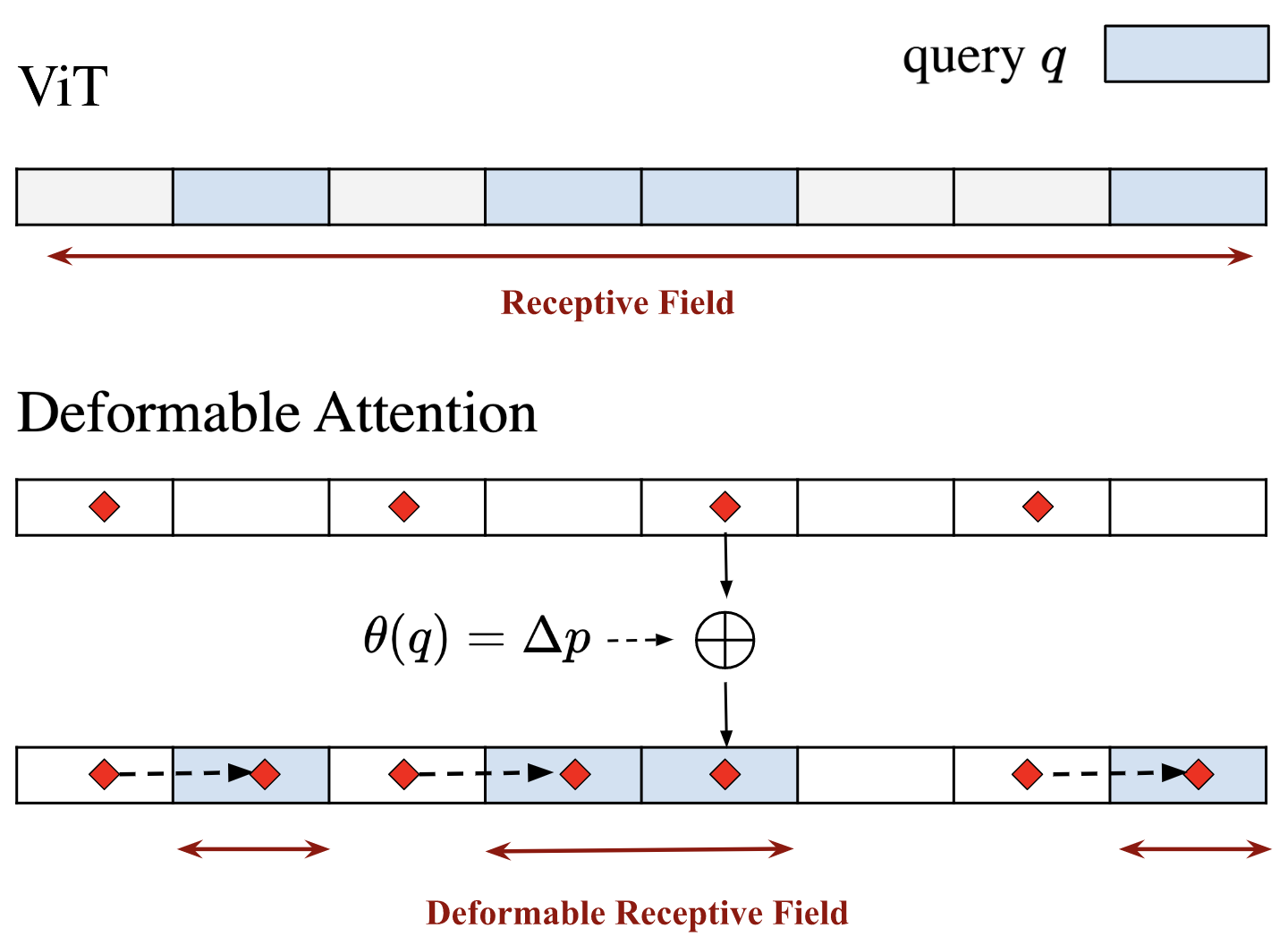}
\vspace{-0.1in}
\caption{\small{Comparison of Deformable Attention (DAT) with standard Vision Transformer (ViT) in our setting using slice's embedding vector. Given a query $q$, ViT pays attention to all possible positions including possibly less relevant feature maps while DAT learns important regions based on grid points (red points) and their shifted vectors using offsets $\Delta p$ predicted by $\theta(q)$.}}
\label{fig:deformable-att}
\end{figure}

To mitigate these problems, we use the deformable self-attention mechanism which is recently introduced in supervised learning such as object detection and image classification \cite{zhu2020deformable,xia2022vision}. Generally this strategy seeks important positions of keys and value pairs in self-attention in a dependent-way rather than a fixed window size as ViT (Figure 3). Specifically, these important regions are learnt using an offset network that takes input query features and returns corresponding offsets whose regions subsequently are used to sample candidates keys/values (Figure \ref{fig:deformable-att}). In this work, we use this deformable attention to SSL for the first time, aiming to learn the association among feature embedding of 2D slices. We call this as Inter Deformable Attention and denote by $f_{\mathrm{inter}}$.
\vspace{-0.1in}
\vspace{-0.1in}
The $f_{\mathrm{inter}}$ contains $N$ identical stacked layers. Each layer is composed of multi-head attention (MHA) layer followed by a simple feed-forward layer. Given an input tensor $Y\in \mathbb{R}^{D \times F_{\mathrm{in}}}$ added with a positional encoding to provide order information, the output of a single head $h$ at each layer using deformable attention can be computed by the following step.
\begin{gather} 
    q^{(h)} = YW_{q}^{h},\ \tilde{k}^{(h)} = \tilde{Y}W_{k}^{h},\  \tilde{v}^{(h)} = \tilde{Y}W_{v}^{h}  \\
    \textrm{with }\tilde{Y} = \phi\left(Y; p + \Delta p\right),\ \Delta p = \theta_{\textrm{offset}}\left(q^{(h)}\right)
\end{gather}
where $W_{q}^{h},\, W_{k}^{h}$ and $W_{v}^{h} \in \mathbb{R}^{F_{\mathrm{in}} \times d_{h}^{v}}$ are learned linear transformation that map the input $Y$ to queries, keys, values respectively; $\theta_{\mathrm{offset}}$ be the offset network that takes input as queries $q^{h}$ and returns the offsets $\Delta p$; $p \in \mathbb{R}^{D_{G}\times 2}$ denotes for the uniform grid of points with $D_G = D/r$ by a factor $r$ to down-sample the grid size; finally $\phi(.;.)$ be a differentiable linear interpolation function used to sample important key/queries pairs inside predicted offsets.

We now compute the output of a deformable attention head $h$ as:
\begin{equation}
    O^{(h)} = \sigma\left(q^{(h)}\tilde{k}^{(h)\top}/\sqrt{d^{(h)}}  + \phi(\hat{B}; R)\right)\tilde{v}^{(h)}
    \label{eq:deformable-att}
\end{equation}
where $\sigma(.)$ denotes the softmax function, $d^{(h)}$ is the dimension of each head $h$, $\hat{B} \in \mathbb{R}^{(2D-1)}$ be a relative position bias matrix, $R$ be the relative position offsets. More details on this bias matrix, we refer the readers to \cite{liu2021swin,xia2022vision}. The outputs of all heads ($\mathrm{MHA}$) are aggregated by concatenating and projecting again as $ \mathrm{MHA} =  \mathrm{Concat}\left[O^{(1)}, ..., O^{(Nh)}\right] W^{O}$
where $W^{O} \in \mathbb{R}^{d_{v} \times d_{v}}$ is another learned linear transformation and $Nh$ is the number of heads.

Given defined $f_{\mathrm{inter}}$, we construct a 3D feature representation $\mathbf{Z}_{s}$ for an augmented view $\mathbf{X}_{s} = \{\mathbf{x}_{1s},\mathbf{x}_{2s},...,\mathbf{x}_{ns}\}$ of $\mathbf{X}$ as follows. We denote by 
\begin{equation}
    \mathbf{Y} = \left[\{f_{\mathrm{intra}}(\mathbf{x}_{1s})\}_{l=1}^{L}, \ldots,\{f_{\mathrm{intra}}(\mathbf{x}_{ns})\}_{l=1}^{L}\right]
    \label{eq:multi-level}
\end{equation}
be the stacked input vectors with $\{f_{\mathrm{intra}}(\mathbf{x}_{is})\}_{l=1}^{L},\ i \in [1,n]$  indicates the multi-level features of image $\mathbf{x}_{is}$ taken from the $L$ last layers in $f_{\mathrm{intra}}$. We then normalize the ouput of $f_{\mathrm{inter}}$ and obtain
\begin{equation}
\scalebox{1.0}{$
    \mathbf{Z_{s}} = 
    f_{\mathrm{inter}}(\mathbf{Y)} / || f_{\mathrm{inter}}(\mathbf{Y)} ||_2
    \label{eq:3D-feature}$}
\end{equation}
which is the holistic feature of $\mathbf{X}_s$. The embedding $\mathbf{Z}_t$ for transformation $t \in T$ is computed analoguously. The clustering agreement for 3D volumes generalized from Eq.(\ref{eq:loss_idea_intra}) can be defined as
\begin{equation}
\scalebox{0.95}{$
    L_{\mathrm{inter}}(\mathbf{Z}_{t}, \mathbf{q}^{3D}_t, \mathbf{Z}_{s}, \mathbf{q}^{3D}_s) = l(\mathbf{Z}_{t}, \mathbf{q}^{3D}_{s}) + l(\mathbf{Z}_{s}, \mathbf{q}^{3D}_{t})$}
\label{eq:loss_idea_inter}
\end{equation}
where $\mathbf{q}^{3D}_{s}, \mathbf{q}^{3D}_{t}$ are codes of $\mathbf{Z}_{s},\mathbf{Z}_{t}$ obtained by solving the matching problem in Eq.\eqref{eq:matching} where inputs are 3D augmented views' feature represents across 3D volumes $\mathbf{X_{i}}$ in a batch size $B \in \mathbb{D}$. Intuitively, two 3D features $\mathbf{Z}_s$ and $\mathbf{Z}_t$ should be identical in their cluster assignments. 
Finally, by minimizing over samples in $\mathbb{D}$, we jointly learn both $f_{\mathrm{intra}}, f_{\mathrm{inter}}$ and $\mathbf{C}$ through
\vspace{-0.1in}
\begin{multline}
    L_{3D} =  \min_{f_{\mathrm{intra}}, \mathbf{C}, f_{\mathrm{inter}}} \mathrm{E}_{\mathbf{X} \in \mathbb{D}}\left[L_{\mathrm{inter}}(\mathbf{Z}_{t}, \mathbf{q}^{3D}_t, \mathbf{Z}_{s}, \mathbf{q}^{3D}_s)\right] \\ \textrm{ with } s,t \sim T.
    \label{eq:3D-cluster}
\end{multline}
\vspace{-0.24in}
\subsection{Masked Feature Embedding Prediction}
To enhance long-term dependence learning of $f_{\mathrm{inter}}$, we additionally introduce a new SSL proxy task inspired by the BERT language model~\cite{devlin2018bert}.
Given a set of 2D slice embedding vectors $\mathbf{Y}$ in Eq.(\ref{eq:multi-level}) obtained from $\mathbf{X_{s}}\ (\mathbf{X} \in \mathbb{D}, s \sim T$), we dynamically mask some inputs 
$\{f_{\mathrm{intra}}(\mathbf{x}_{is})\}_{l=1}^{L}$, $i \in  [1,n]$
and ask the \textit{Inter Deformable Attention} to predict missing encoding vectors given the unmasked embedding vectors. To do this, we define a binary vector $\mathbf{m} = (m_1,\ldots, m_n)$ of length $n$ where $m_i = 1$ indicate the input $i$-th of $\mathbf{Y}$ will be masked and $0$ otherwise. The input for SSL task then is defined as:
\begin{equation}
\scalebox{1.0}{$
    \mathbf{m} \odot \mathbf{Y} = \left\{\begin{matrix}
[\mathrm{MASK}], \ m_i = 1 \\ 
\{f_{\mathrm{intra}}({\mathbf{x}_{is}})\}_{l=1}^{L}, \ m_i = 0
\end{matrix}\right.$}
\end{equation}
where $\mathrm{MASK}$ is a learnable parameter during the training step.
We denote by $f_{\mathrm{decode}}$, a fully connected layer, that takes the outputs of  $f_{\mathrm{inter}}$ and predicts masked vectors. For each $\mathbf{m}$, we randomly assign $m_i = 1$ for $10\%$ of $\mathbf{m}$.
The output of $f_{\mathrm{decode}}$ at each masked $\mathbf{y}_i$ is:
\begin{equation}
    \mathbf{y}_i = \mathbf{W_{d}}\mathbf{h}_{i}^{N} + \mathbf{b}_{i},\ \textrm{ where } m_i = 1.
\end{equation}
with $\mathbf{W}_{d} \in \mathbb{R}^{F_{in} \times F_{D}}$ and $\mathbf{b}_{i} \in \mathbb{R}^{F_{in}}$ are fully-connected layers and biases respectively.
The masked feature embedding prediction is defined as:
\vspace{-0.1in}
\begin{multline}
    L_{mask} =\hspace{-0.1cm} \footnotesize \min\limits_{\substack{f_{\mathrm{inter}}\\ f_{\mathrm{decode}}}} \mathrm{E}_{\substack{\,\mathbf{X} \in \mathbb{D}\\ s \sim T}} \left[\sum\limits_{i:m_i = 1} || f_{\mathrm{intra}}({\mathbf{x}_{is}}) \right. \\ - \left. f_{\mathrm{decode}}(f_{\mathrm{inter}}(\mathbf{m} \odot \mathbf{Y})) ||_2 \vphantom{\frac12}\right]
    \label{eq:mask-predict}
\end{multline}

\section{Experiment Results}
\subsection{Data and Baseline Setup}
\paragraph{\textbf{Pre-training and Downstream Tasks}}
We describe the details of datasets used for pre-training and downstream tasks in Table~\ref{tab:pre-training} and Table~\ref{tab:downstream}, respectively. In summary, there are thirteen datasets comprising LUNA2016~\cite{setio2015automatic}, LiTS2017~\cite{bilic2019liver}, BraTS2018~\cite{bakas2018identifying}, MSD (Heart) \cite{simpson2019large}, MOTS \cite{zhang2021dodnet}, LIDC-IDRI \cite{clark2013cancer, armato2011lung}, RibFrac \cite{jin2020deep}, TCIA-CT \cite{clark2013cancer, harmon2020artificial}, NIH ChestX-ray8 \cite{wang2017chestx}, MMWHS-CT/MMWHS-MRI \cite{zhuang2016multi}, VinDR-CXR \cite{nguyen2022vindr}, and JSRT \cite{shiraishi2000development,van2006segmentation}. In pre-training settings, we mainly evaluate in two scenarios, namely \textit{Universal} and \textit{Unified} following prior works of \citet{DBLP:journals/corr/abs-2010-06107} and \citet{xie2021unified}, respectively. However, we cannot access the dataset called ``Tianchi dataset'' in \textit{Unified} setting thus we only train with five remaining datasets. The downstream tasks are conducted in three contexts with diverse applications as described Table~\ref{tab:downstream}. For objective assessment, we use Intersection over Union (IoU) computed on 3D data for segmentation, Area Under the Curve (AUC) for 3D classification, Dice coefficient scores for 2D segmentation, and Average Precision with IoU=$0.5$ for multi-object detection. 
\vspace{-0.1in}
\begin{table}[H]
\caption{\small{Overview pre-training settings in our experiment. The \textit{Universal} setting uses four unlabeled 3D datasets while \textit{Unified}} uses six unlabeled datasets including mixed 2D and 3D modalities.}
\vspace{-0.2in}
\begin{center}
\scalebox{0.7}{
\begin{tabular}{ccccc}
\toprule
\textbf{Setting }    & \textbf{Pre-Training Data}  & \textbf{Modality}     & \textbf{Num}    & \textbf{Access}   \\
\midrule
\multirow{4}{*}{Universal} & LUNA2016       & 3D CT    & 623    &   \checkmark \\
       & LiTS2017       & 3D CT    & 111    &    \checkmark \\
       & BraTS2018      & 3D MRI   & 760    &    \checkmark \\
       & MSD (Heart)    & 3D MRI   & 30     &   \checkmark \\
       \midrule
\multirow{6}{*}{Unified}   & MOTS           & 3D CT    & 936    &  \checkmark \\
       & LIDC-IDRI      & 3D CT    & 1008   &  \checkmark \\
       & Tianchi        & 3D CT    & N/A    &  $\times$          \\
       & RibFrac        & 3D CT    & 420    &  \checkmark  \\
       & TCIA-CT        & 3D CT    & 1300   &  \checkmark  \\
       & \multicolumn{1}{l}{NIH ChestX-ray8} & \multicolumn{1}{l}{2D Xrays} & \multicolumn{1}{l}{108948} &  \checkmark \\
       \bottomrule
\end{tabular}}
\label{tab:pre-training}
\end{center}
\end{table}
\vspace{-0.2in}
\begin{table}[H]
\caption{\small{Overview downstream tasks used in our experiment. \textbf{Seen Domain} indicates for downstream tasks where the training data was used in the pre-training step without labels, \textbf{Unseen Domain} means that datasets in pre-training and downstream are different.}}
\vspace{-0.2in}
\begin{center}
\scalebox{0.65}{
\setlength\tabcolsep{1.5pt}
\begin{tabular}{cccccc}
\toprule
\textbf{Setting}  & \textbf{Testing Data}      & \textbf{Modality} & \textbf{Num} & \textbf{Pre-training} & \textbf{Task}  \\
\midrule
\multirow{2}{*}{\begin{tabular}[c]{@{}c@{}}{Seen domain}\\ {in Universal}\end{tabular}}  
    & BraTS2018   & 3D MRI   & 285  & Universal  & {Tumor Segmentation} \\
    & LUNA 2016   & 3D CT    & 888     & Universal   & Lung Nodes Classification \\ 
\midrule
\multirow{3}{*}{\begin{tabular}[c]{@{}c@{}} {Unseen Domain}\\ {in Universal}\end{tabular}}
    & MMWHS-CT    & 3D CT    & 20   & Universal   & Heart Structures Segmentation    \\
    & MMWHS-MRI   & 3D MRI   & 20   & Universal   & Heart Structures Segmentation    \\
    & VinDR-CXR   & 2D X-ray & 4394 & Universal   & Abnormal Chest Detetction \\
\midrule
\begin{tabular}[c]{@{}c@{}}{Unseen Domain}\\ {in Unified}\end{tabular} 
    & JSRT  & 2D X-ray   & 247    & Unified   & Multi-Organ Segmentation  \\
\bottomrule
\end{tabular}} 
\end{center}
\label{tab:downstream}
\end{table}

\paragraph{\textbf{Competing Algorithms}}
We implement variations of Deepcluster and SwAV based the proposed method and compare with the following approaches:
\begin{itemize}
    \item \textit{2D SSL methods}: SimCLR \cite{chen2020simple}, PGL \cite{xie2020pgl}, Moco-v2~\cite{chen2020improved}, Deep-Cluster-v2 \cite{caron2020unsupervised}, SwAV \cite{caron2020unsupervised}, Barlow-Twins \cite{zbontar2021barlow}, Moco-V3 \cite{chen2021empirical}, PCRL \cite{zhou2021preservational}, and DINO \cite{caron2021emerging}. Both Moco-v3 and DINO use Pyramid Transformer Unet \cite{xie2021unified} as backbone.
    \item \textit{3D SSL methods}: 3D Rotation, 3D JigSaw \cite{taleb20203d}, Universal Model~\cite{DBLP:journals/corr/abs-2010-06107}, Models Genesis \cite{zhou2021models}, TransVW \cite{haghighi2021transferable}, SwinViT3D \cite{tang2022self}, and our two implementations for the 3D case of Deepcluster-v2 and SwAV, namely 3D-Deepcluster and 3D-SwAV.
    \item \textit{2D/3D supervised pre-trained methods}: 2D pre-trained ImageNet \cite{he2016deep}, I3D \cite{carreira2017quo}, NiftyNet \cite{gibson2018niftynet},  and Med3D \cite{chen2019med3d}.
    \item \textit{Other methods}: training from scratch for 2D or 3D using ResNet-50, V-Net architecture \cite{milletari2016v},  3D-Transformer \cite{hatamizadeh2022unetr}, Pyramid Transformer Unet (PTU) \cite{xie2021unified} and finally USST \cite{xie2021unified},  a joint 2D and 3D approach similar to ours.
\end{itemize}

Most baseline results are taken from \cite{DBLP:journals/corr/abs-2010-06107} and \cite{xie2021unified}. With LUNA2016 dataset, we use the latest ground-truth, denoted as LUNA2016-v2, and provide results obtained when training with batch sizes of $8, 16, 32$, each with two trial times. 
For new competitors, we describe experiment setups in the appendix. In short, for 2D self-supervised methods (ResNet-50 backbone) such as Moco-v2 or Barlow-Twins, we extract all 2D slices from 3D volumes in pre-training data and train SSL tasks with $100$ epochs. 
With state-of-the-art 3D SSL methods TransVW and SwinViT3D, we download pre-trained weights and use published implementation to fine-tune as author's suggestions. For two our implementations of 3D-Deepcluster and 3D-SwAV, we train with all 3D data of \textit{Universal} in pre-training step. 

\vspace{-0.1in}
\begin{table}[!hbt]
\caption{\small{Comparing SSL approaches on \textbf{Seen Domains} trained on the \textit{Universal setting}. Two top results in 2D or combined 2D-3D SSL data are red, blue. The best values in 3D-based methods and overall are in bold and underlined respectively. N/A indicates pre-trained models that are unable to transfer (Universal Model's results are not available in LUNA2016-v2).}}
\vspace{-0.2in}
\label{tab:seen-domain-univer}
\setlength\tabcolsep{2.0pt}
\begin{center}
\scalebox{0.6}{
\begin{tabular}{cccc}
\toprule
\multicolumn{1}{c}{\textbf{Pre-training}} & \textbf{Method} & \begin{tabular}[c]{@{}c@{}}\textbf{BraTS2018} \\ \textbf{(MRI - Segmentation)}\end{tabular} & \begin{tabular}[c]{@{}c@{}}\textbf{LUNA2016-v2} \\ \textbf{(CT - Classification)}\end{tabular}\\
\midrule
        & Scratch (3D) & 58.51 $\pm$ 2.61 & 94.15 $\pm$ 3.97\\
        N/A & V-Net &  59.01 $\pm$ 2.59 & 95.85 $\pm$ 1.09 \\
        & 3D-Transformer & 66.54 $\pm$ 0.40 & 85.15 $\pm$ 2.62 \\
\hdashline
        & I3D & 67.83 $\pm$ 0.75 & 92.43 $\pm 2.63$ \\
        3D Supervised & NiftyNet  & 60.78 $\pm$ 1.60 & 94.16 $\pm 1.52$\\
        & Med3D & 66.09 $\pm$ 1.35 & 91.32 $\pm$ 1.47 \\  \hdashline
\multicolumn{1}{c}{\multirow{8}{*}{3D Self-supervised}}
        & 3D-Rotation  & 56.48 $\pm$ 1.78 & \textbf{\underline{95.91} $\pm$ 1.26} \\
        & 3D-JigSaw & 59.65 $\pm$ 0.81 & 89.12 $\pm$ 1.71 \\
        & Models Genesis  & 67.96 $\pm$ 1.29 & 92.46 $\pm$ 5.54 \\
        & Universal Model & \textbf{72.10 $\pm$ 0.67} & N/A \\
        & 3D-DeepCluster & 59.20 $\pm$ 1.69 &  89.03 $\pm$ 2.56 \\
        & 3D-SwAV & 62.81 $\pm$ 1.03 & 88.79 $\pm$ 5.48 \\
        & TransVW & 68.82 $\pm$ 0.38 & 93.84 $\pm$ 6.73 \\
        & SwinViT3D & 70.58 $\pm$ 1.27 & 88.68 $\pm$ 2.63 \\
\Xhline{2\arrayrulewidth}
N/A & Scratch (2D) & 66.82 $\pm$ 1.32  & N/A \\
\hdashline
2D Supervised & Pre-trained ImageNet & {71.24 $\pm$ 2.30} & N/A \\
\hdashline
		& SimCLR  & 70.37 $\pm$ 1.11 & N/A \\
		& Moco-v2  & 70.82 $\pm$ 0.22 & N/A \\
        2D Self-supervised & Barlow-Twins  & 67.35 $\pm$ 0.55 & N/A     \\
        & Deep-Cluster-v2  & 69.21 $\pm$ 2.10 & N/A      \\
		& SwAV  & 69.83 $\pm$ 2.44  & N/A \\
\Xhline{2\arrayrulewidth}
\multicolumn{1}{c}{\multirow{2}{*}{2D $\&$ 3D Self-supervised}}
		& \textbf{Our (Deep-Cluster-v2)} & {\color[HTML]{3531FF} {72.81 $\pm$ 0.15}} & {\color[HTML]{3531FF} {93.91 $\pm$ 0.67}} \\
		& \textbf{Our (SwAV)} & {\color[HTML]{FF0000} {\underline{73.03} $\pm$ 0.42}} & {\color[HTML]{FF0000} {94.22 $\pm$ 1.11}} \\
\bottomrule
\end{tabular}}
\end{center}
\label{tab:see-domains}
\end{table}
\vspace{-0.2in}
\subsection{Implementation Details}
\textbf{Pre-training} Our method is trained in three stages. Stage 1 learns $f_{\mathrm{intra}}$ using Eq.\,(\ref{eq:2D-clustering}) with $100$ epochs using batch size of $1024$ images, Stage 2 learns $f_{\mathrm{inter}}$ using Eq.\,(\ref{eq:mask-predict}) with $100$ epochs using batch size of $12$ volumes, and Stage 3 learns for both $f_{\mathrm{intra}}, f_{\mathrm{inter}}$ using Eq.\,(\ref{eq:3D-cluster}) also with $100$ epochs and batch size of $12$ volumes.

\begin{table}[H]
\caption{\small{Comparing SSL approaches on \textbf{Unseen Domains} trained on the \textit{Universal setting}. Two top results in 2D or combined 2D-3D SSL data are red, blue. The best values in 3D-based methods and overall are in bold and underlined respectively}. N/A indicates for pre-trained models that are unable to transfer.}
\vspace{-0.2in}
\setlength\tabcolsep{2.0pt}
\label{tab:unseen-domain-univer}
\begin{center}
\scalebox{0.6}{
\begin{tabular}{c c c c c }
\toprule
\multicolumn{1}{c}{\textbf{Pre-training}} & \textbf{Method} & \begin{tabular}[c]{@{}c@{}}\textbf{MMWHS} \\ \textbf{(CT - Segm.)}\end{tabular} & \begin{tabular}[c]{@{}c@{}}\textbf{MMWHS} \\ \textbf{(MRI - Segm.)}\end{tabular} & \begin{tabular}[c]{@{}c@{}}\textbf{VinDr-CXR} \\ \textbf{(X-ray - Detect.)}\end{tabular}\\
\midrule  
& Scratch (3D) & 68.29 $\pm$ 1.68 & 67.04 $\pm$ 2.18 & N/A \\
N/A & V-Net  & 69.66 $\pm$ 3.65 & 67.50 $\pm$ 3.76 & N/A \\
& 3D-Transformer  & 67.30 $\pm$ 2.29 & 67.64 $\pm$ 2.21 & N/A \\
\hdashline
		& I3D  & 76.63 $\pm$ 2.32 & 66.71 $\pm$ 1.27 & N/A \\
		3D Supervised & Nifty Net  & 74.91 $\pm$ 2.78 & 64.60 $\pm$ 1.96 & N/A \\
        & Med3D  & 75.01 $\pm$ 0.74 & 63.43 $\pm$ 0.61 & N/A \\  \hdashline 
		\multicolumn{1}{c}{\multirow{8}{*}{3D Self-supervised}} & 3D Rotation & 67.54 $\pm$ 2.80 & 71.36 $\pm$ 1.70 & N/A \\
		& 3D Jigsaw  & 68.40 $\pm$ 2.92 & 72.99 $\pm$ 2.54 & N/A \\
		& Model Geneis  & 76.48 $\pm$ 2.89 & 74.53 $\pm$ 1.69 & N/A \\
        & Universal Model & 78.14 $\pm$ 0.77 & 77.52 $\pm$ 0.50 & N/A \\
        & 3D-DeepCluster  & 69.47 $\pm$ 1.44 & 75.83 $\pm$ 2.29 & N/A\\
        & 3D-SwAV  & 69.90 $\pm$ 1.31 & 69.41 $\pm$ 1.93 & N/A\\
        & TransVW  & \textbf{79.74 $\pm$ 2.78} & 75.08 $\pm$ 2.04 & N/A\\
        & SwinViT3D  & 70.19 $\pm$ 1.23  & \textbf{78.25 $\pm$ 1.66} & N/A\\
\Xhline{2\arrayrulewidth}
\multicolumn{1}{c}{N/A}  & Scratch (2D) & 74.25 $\pm$ 2.05  & 52.34 $\pm$ 4.31 & 24.35 $\pm$ 0.04 \\
\hdashline
\multicolumn{1}{c}{2D Supervised} & Pre-trained ImageNet & 73.49 $\pm$ 3.15 & {\color[HTML]{000000} 72.66 $\pm$ 2.46} & {27.82 $\pm$ 0.29} \\
\hdashline 
		& SimCLR  & 78.56 $\pm$ 2.12 & 72.72 $\pm$ 1.29 & 26.87 $\pm$ 0.32  \\
		& Moco-v2  & {80.25$\pm$ 0.93} & 71.85 $\pm$ 1.25 & 27.20 $\pm$ 0.66 \\
		2D Self-supervised & Barlow-Twins & 80.95 $\pm$ 2.47 & 70.90 $\pm$ 1.89 & 26.83 $\pm$ 0.13 \\
		& Deep-Cluster-v2  & 81.03 $\pm$ 1.17 & 74.51 $\pm$ 1.92 & \color[HTML]{0000FF}{28.03 $\pm$ 0.41} \\
		& SwAV  & 82.15 $\pm$ 1.19 & {{{74.50 $\pm$1.20}}} & 27.70 $\pm$ 0.22 \\ \Xhline{2\arrayrulewidth}
\multicolumn{1}{c}{\multirow{2}{*}{\rotatebox[origin=c]{0}{2D $\&$ 3D Self-supervised}}}
		& \textbf{Our (Deep-Cluster-v2)} & {\color[HTML]{0000FF} {83.58 $\pm$ 1.54}} & {\color[HTML]{0000FF} {78.14 $\pm$ 1.32}} & \color[HTML]{FF0000}{\underline{28.47} $\pm$ 0.40} \\
		& \textbf{Our (SwAV)} & {\color[HTML]{FF0000} {\underline{84.89} $\pm$ 0.68}} & {\color[HTML]{FF0000} {\underline{78.73} $\pm$ 1.21}} & 27.47 $\pm$ 0.18 \\
\bottomrule
\end{tabular}}
\end{center}
\label{tab:unseen}
\end{table}
\vspace{-0.25in}
\begin{table}[H]
\caption{\small{Performance comparison on the 2D JSRT segmentation tasks using different SSL approaches trained on the \textit{Unified setting}. Two top results are illustrated in red and blue respectively.}}
\vspace{-0.2in}
\setlength\tabcolsep{2.0pt}
\begin{center}{
\scalebox{0.65}{
\begin{tabular}{ccclll}
\toprule
\multirow{2}{*}{\textbf{Pre-training}} & \multirow{2}{*}{\textbf{Methods}} & \multirow{2}{*}{\textbf{Backbone}} & \multicolumn{3}{c}{\textbf{JSRT (X-ray, seg.)}} \\
&&& 20\%  & 40\%  & 100\%  \\
\midrule
\multicolumn{1}{c}{\multirow{2}{*}{N/A}}
& Scratch CNN& ResNet-50& \multicolumn{1}{c}{84.05} & \multicolumn{1}{c}{87.63} & \multicolumn{1}{c}{90.96} \\ 
& Scratch PTU & Transformer & \multicolumn{1}{c}{85.55} & \multicolumn{1}{c}{88.83} & \multicolumn{1}{c}{91.22} \\ \hdashline
2D Supervised& Pre-trained ImNet  & ResNet-50& 87.90 & 90.01 & 91.73 \\ \hdashline
& Moco-v2 & ResNet-50& 88.65 & 91.03 & 92.32 \\
& PGL  & ResNet-50& 89.01 & 91.39 & 92.76 \\
2D Self-Supervised  & PCRL & ResNet-50& 89.55 & 91.53 & 93.07 \\
& Moco-v3 & Transformer & 90.07 & 91.75 & 92.68 \\
& DINO & Transformer & 90.40 & 92.16 & 93.03 \\  \hline
& USST & Transformer & {\color[HTML]{FF0000}{91.88}} & {\color[HTML]{FF0000}{93.15}}& 94.08 \\
2D \& 3D Self-supervised & \textbf{Our (DeepCluster-V2)} & ResNet-50& \color[HTML]{0000FF}{90.60}& 92.87 & \color[HTML]{0000FF}{94.31}  \\
& \textbf{Our (SwAV)} & ResNet-50& 89.98& \color[HTML]{0000FF}{93.03}& {\color[HTML]{FF0000}{94.45}}  \\
\bottomrule
\end{tabular}}}
\end{center}
\label{tab:unseen-unified}
\end{table}
\vspace{-0.1in}
We use ResNet-50 as the backbone for 2D feature extractor $f_{\mathrm{intra}}$. The features for each image are concatenated from five blocks of ResNet-50. The architecture of $f_{\textrm{inter}}$ has four pyramid structure blocks composed from deformable attention (Eq.~\eqref{eq:deformable-att}).
Details for these configurations can be found in Appendix. In the \textit{Universal} or \textit{Unified} setting, we utilize all 3D data as benchmarks and further extract 2D slices from them to train $f_{\mathrm{intra}}$ in Stage 1. All experiments are conducted on a A100-GPU system with 4 GPUs, 40GB of memory each with Pytorch. It takes in average $30$ hours to finish the pre-training step.

\paragraph{Downstream Task} we use the SGD with a learning rate selected in a set $\{0.1,\,0.01\}$ and select a specific number of epoch depended on downstream task properties (Appendix).
The results are reported by running training-testing five times and computing the average values (except LUNA2016-v2 dataset). For the 2D/3D segmentation task, we use the pre-trained 2D-CNN feature extractor in each 2D baseline ($f_{\mathrm{intra}}$ in our method) as the network backbone of a 2D U-net~\cite{ronneberger2015u}. This network is trained with cross-entropy and dice loss. We predict segmentation at each 2D slice and merge results for 3D volumes. The 3D classification is solved by building on top of the deformable transformer two fully-connected layers and fine-tuning for both $f_{\mathrm{inter}}$ and $f_{\mathrm{intra}}$ with the cross-entropy loss. For the 2D object detection task (VinDr-CXR), we use the 2D-CNN feature extractor ($f_{\textrm{intra}}$) as the backbone of Faster R-CNN model \cite{ren2015faster}.

\subsection{Performance Evaluation}
\paragraph{\textbf{Dimension-specific vs. Cross-dimension Pre-training}}
Tables \ref{tab:seen-domain-univer} and \ref{tab:unseen-domain-univer} indicate that 2D CNN based-models cannot transfer to the 3D lung node classification task in LUNA2016-v2 (denoted N/A) given input 3D volumes. Likewise, due to data compatibility issues, 3D CNN-based methods cannot apply for abnormal chest detection in X-rays. In contrast, our models pre-trained on several medical datasets can be transferred successfully in both cases due to the hybrid CNN-Transformer architecture. We argue that such property is one of the most valuable points of this study.

As compared with plain 2D-SwAV, Deepcluster-V2, and their extended versions with 3D CNN, namely 3D-SwAV and 3D-Deepcluster, we show a significant improvement in several settings, especially for segmentation tasks (Tables \ref{tab:seen-domain-univer},\ref{tab:unseen-domain-univer}). For instance, a gain performance of 2-3$\%$ on average on BraTS, MMWHS-CT/MRI datasets. Furthermore, we also achieve better accuracy on 3D classification and 2D object detection, although with smaller margins. In conclusion, this analysis shows that exploiting deformable self-attention in conjunction with 2D CNN to model 3D volume features in our framework is a promising approach.

\paragraph{\textbf{Comparison to SOTA Methods and Visualizations}}
In the Universal setting, except the LUNA2016-v2 case where we are third, our methods based on Deepcluster-V2 or SwAV hold the best records on BraTS, MMWHS-CT/MRI segmentation tasks compared with remaining baselines, especially with cutting edges 3D-SSL methods as Universal Model, TransVW or SwinViT3D (using Swin Transformer). With the VinDr-CXR detection task, we continue to reach the best rank, followed by the plane 2D Deepcluster-v2 though with smaller margins. In the Unified setting, we also surpass competitors (100$\%$ data), especially with USST, a method using Pyramid Vision Transformer trained on mixed 2D and 3D data. However,  USST works better than us when decreasing training data to $40\%$ and $20\%$. We consider this as a potential limitation that needs to improve. Though it's worth noting that we could not access all data as USST in the pre-training step, as shown in Table \ref{tab:pre-training}.

\begin{figure}[H]
    \centering
    \includegraphics[width=0.45\textwidth]{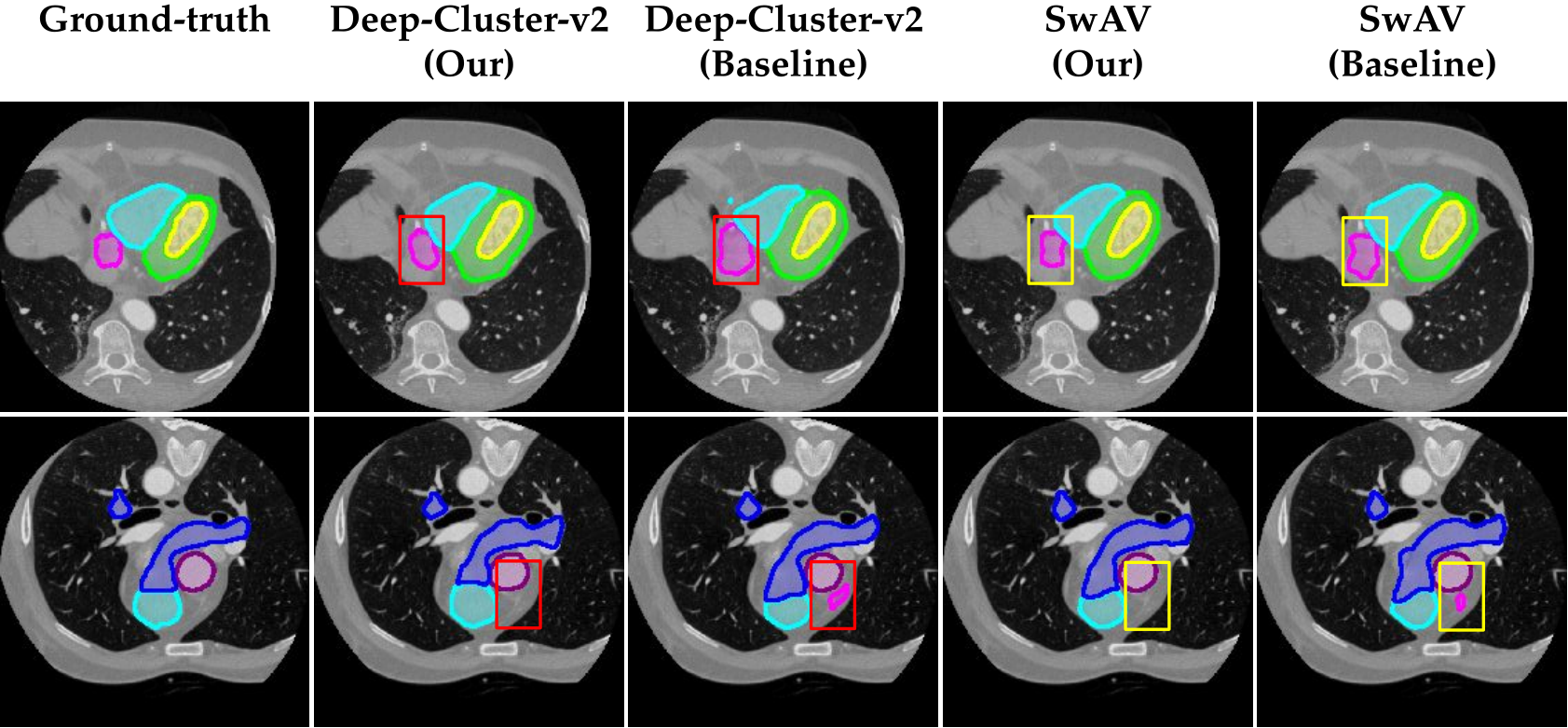}
    \caption{\small{Heart structure segmentation on MMWHS-CT. The figures show that baselines tend to over-segment in the first row while generating noise regions in the second row. On the contrary, our methods produce more precise results.}}
    \label{fig:comparison}
\end{figure}

For visualization results, we provide a typical example of multi-modal heart segmentation for MMWHS-CT in Figure \ref{fig:comparison} and abnormal Chest X-ray detection in Figure \ref{fig:vin-dr-compare}. More examples can be found in the Appendix.

\paragraph{\textbf{Computational Complexity and Ablation Study}} 
We compare the total parameters with top baselines and methods using Transformer in Table \ref{tab:memory}. In short, our total parameter is half of the SwinViT3D but we attain better performance in overall. The contributions of proposed SSL tasks and multi-level features are presented in Table \ref{tab:ablation}, where all components contribute to overall accurate growth.
\vspace{-0.2in}
\begin{figure}[H]
    \centering
    \includegraphics[width=0.39\textwidth]{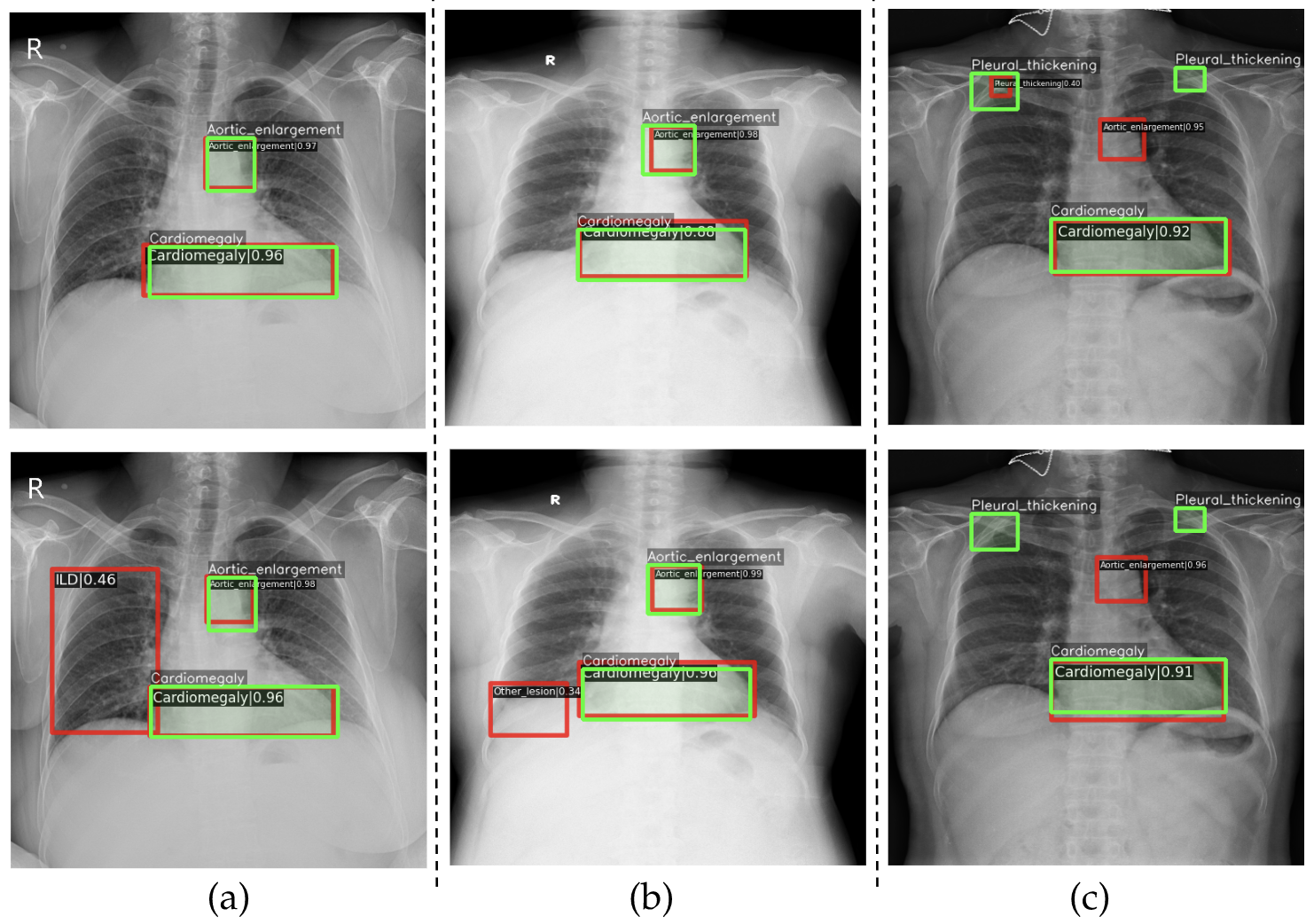}
    \vspace{-0.1in}
    \caption{\small{Visualization of our results based on DeepCluster-V2 (top row) and pre-trained ImageNet (second row) in abnormal Chest X-ray. Green and red indicates for ground-truths and predictions.}}
    \label{fig:vin-dr-compare}
\end{figure}
\vspace{-0.2in}
\begin{table}[H]
\setlength\tabcolsep{6pt}
\caption{\small{Ablation studies for the SwAV on heart segmentation.}}
\vspace{-0.2in}
\begin{center}
\scalebox{0.72}{
\begin{tabular}{ccc}
\toprule
\textbf{Setting} & \textbf{MMWHS - CT} & \textbf{MMWHS - MRI}\\
\midrule
W/o mask prediction & 82.53 & 77.35 \\
W/o 3D clustering & 81.97 & 76.18 \\ 
Full model & 84.89 & 78.73 \\ 
Full model w/o multi-feature & 83.56 & 78.12\\
\bottomrule
\end{tabular}}
\label{tab:ablation}
\end{center}
\end{table}
\vspace{-0.2in}
\begin{table}[H]
\setlength\tabcolsep{5pt}
\caption{\small{Computational complexity of top baselines and transformer-based methods. For USST, we follow general descriptions in paper to re-configure architecture}}.
\vspace{-0.2in}
\begin{center}
\scalebox{0.75}{
\begin{tabular}{cccccc}
\toprule
 & SwinViT3D  & TransVW & Universal Model & USST & Our  \\
\midrule
\textbf{\#Param} & 62.19\,M & 19.7\,M & 19.7\,M & 47.8\,M & 31.16\,M\\
\bottomrule
\end{tabular}}
\label{tab:memory}
\end{center}
\end{table}
\vspace{-0.15in}
\section{Conclusion}
We contribute to the self-supervised learning medical imaging literature a new approach that is efficient in using numerous unlabeled data types and be flexible with data dimension barriers in downstream tasks. To that end, we developed a deformable self-attention mechanism on top of a 2D CNN architecture, which leads to both intra- and inter-correlations formed in our framework. Furthermore, our two novel SSL tasks including 3D agreement clustering and masked embedding predictions impose a tighter constraint in learning feature space, advancing pre-trained network performance in a variety of medical tasks. In the future, we will investigate this method for various SSL approaches, aiming to validate its universality and robustness in real-life medical usage.
\section{Acknowledgement}
This research has been supported by the pAItient project
(BMG, 2520DAT0P2), Ophthalmo-AI project (BMBF, 16SV8639)
 and the Endowed Chair of Applied Artificial Intelligence, Oldenburg University. Binh T. Nguyen wants to thank AISIA Lab, University of Science, and Vietnam National University in Ho Chi Minh City for their support. 
\section{Ethics Statement}
Because our system does not require labeled data and benefits from as much data as possible, the temptation to exploit patient data without their consents arises. Ethical issues may also occur as a result of data collection bias which leads to unfavorable results for ethnically and economically disadvantaged subpopulations, thereby exacerbating the gaps currently existing in healthcare systems.
\begin{small}
\bibliography{aaai23}
\end{small}
\clearpage
\begin{strip}
    \centering
    \LARGE \bf
    Appendix -- Joint Self-Supervised Image-Volume Representation Learning \\ with Intra-Inter Contrastive Clustering
\end{strip}
In this supplement material, we present more information on our deformable self-attention architecture, pre-training, downstream setup, an ablation study when changing the rate of training samples in fine-tuning phases, additional visualization illustrations, and finally, discuss current limitations and future directions for further improvements.

\section{Deformable Self-attention Architecture}

\begin{figure*}[!htb]
\centering
\includegraphics[width=0.78\textwidth]{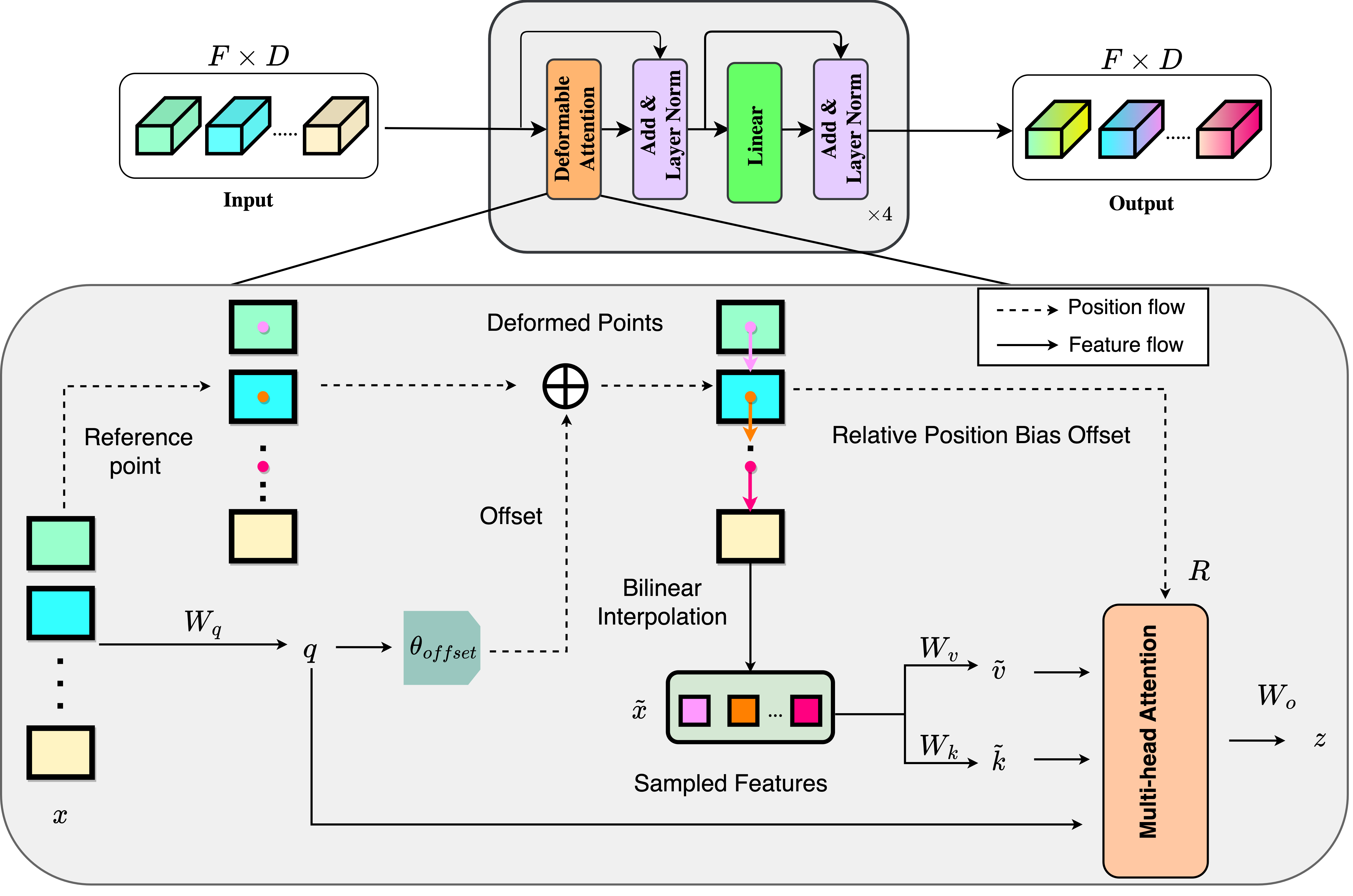}
\caption{\small{The illustration of the Deformable Attention Transformer. 
$F$ refers to the number of frames of our 3D input and $D$ represents 
the embedding dimension for each frame. In our setting, we choose $F = 64$,  $D = 640$.}}
\label{fig:DeTrans}
\end{figure*}

    
To construct a hierarchical feature representation, we build the deformable self-attention with four consecutive blocks as Figure \ref{fig:DeTrans}. Each block involves $6$ number of heads and $6$ offset groups \cite{zhu2020deformable,xia2022vision}. 
Intuitively, this design permits employing a multi-scale information of consecutive 2D slices in 3D volumes, thereby enhancing representation capability while saving computational costs, especially for 3D classification task when attention is estimated for selected local regions.

\section{Pre-Training Setting} 
\paragraph{2D Self-Supervised Baselines:}
We employ VISSL\footnote{\url{https://github.com/facebookresearch/vissl}} to perform pre-training for all 2D self-supervised methods based on ResNet-50 backbone, including SimCLR, Moco-v2,  Barlow-Twins, Deepcluster-v2, and SwAV. All of them are trained with $100$ epochs, batch size of $1024$ images distributed on 4 GPUs of the A100 system. Other parameters are set as default by VISSL. Training data for these 2D-SSL approaches is taken by extracting 2D slices from all 3D volumes in distinct datasets in the \textit{Universal} setting. The optimizes and learning rates are followed as standard 2D Deepcluster-v2 and SwAV.

\paragraph{3D Self-Supervised Baselines:}
For two 3D SSL methods, 3D-Deepcluster and 3D-SwAV, we implemented the 3D variant of Deepcluster-v2 and SwAV by using 3D-CNN, then we trained them with all 3D data in the \textit{Universal} setting using a batch size of $12$ volumes distributed on 4 GPUs of A100 system with $100$ epochs. The optimizer, learning rate, and other configurations are the same as in the 2D case.

\paragraph{Our Method:}
We optimize the proposed method in three stages. Stage 1 learns $f_{\textrm{intra}}$ with ResNet-50 using Eq. (5) with $100$ epochs, i.e., training default Deepcluster-v2 or SwAV. Stage 2 learns $f_{\textrm{inter}}$ and $f_{\textrm{decode}}$ using Eq.(15) with $100$ epochs given a batch size of $12$ volumes. Finally, we jointly optimize $f_{\textrm{intra}}$ and $f_{\textrm{inter}}$ using the novel 3D agreement clustering problem defined in Eq. (12) with $100/200$ epochs in Universal/Unified settings. Due to the expensive computational costs inside the Deepcluster-v2 compared to SwAV, we specified a batch size of $8$ for this method to avoid out-of-memory issues. With SwAV, we used a batch size of $24$ and selected the number of hidden clusters $H = 3000$. 

We leverage both 2D images and 3D data. In the \textit{Universal} setting, since the data is only available in 3D formats, we get all 2D slices from them to train $f_{\textrm{intra}}$ in Stage 1 as other 2D SSL baselines. Afterward, Stages 2 and 3 use 3D data as usual. Similarly, in the \textit{Unified} setting, we also get 2D images from 3D CT and merge them with 2D X-rays in the NIH ChestX-ray8 dataset to train Stage 1. Similarly, stages 2 and 3 only use all available 3D CT as 3D SSL methods. 




\section{Dataset Description and Other Settings in Downstream Tasks}
\subsection{\textbf{Downstream Dataset}}
We briefly provide information on datasets used in downstream task in Table \ref{tab:downstream-dataset}. This includes training, validation, testing sizes, and the employed loss functions in corresponding tasks. Below we describe more details each dataset's properties:
\begin{itemize}
    \item \textit{BraTS2018} \cite{bakas2018identifying}: This dataset comprises magnetic resonance imaging (MRI) volumes of $285$ brain tumor patients. Each participant was scanned using four distinct modalities: T1-weighted, T1-weighted with contrast enhancement, T2-weighted, and T2 fluid-attenuated inversion recovery (FLAIR). The voxel-level labels of the "whole tumor", "tumor core", and "enhancing tumor" are annotated for each patient. Following settings of baselines in \cite{DBLP:journals/corr/abs-2010-06107}, we choose FLAIR images and build model for the "whole tumor". The training and testing rates are indicated in Table \ref{tab:downstream-dataset}. 
    \item \textit{LUNA2016} \cite{setio2015automatic}: This dataset consists of 888 computed tomography (CT) scans, aiming for nodule detection and false positive reduction task. We employed the extended set including 754975 candidates extracted from 888 CT scans, which is 203910 candidates more than the original set (denoted as LUNA2016-v2). The training set has $817$ positive lung nodules out of $377138$ detection candidates. There are $225475$ candidates in the test set, including $459$ positive lung nodules. The average shape for each candidate is $64\times64\times32$.
    
    \item \textit{MMWHS-CT/MRI} \cite{zhuang2016multi}: This dataset is made up of unpaired 20 MRI and 20 CT volumes that span the whole heart substructures and includes seven labeled structures. Following baseline settings \cite{DBLP:journals/corr/abs-2010-06107}, we segment the left atrial blood cavity regions on the CT and MR formats. 
    
    \item \textit{VinDr-CXR} \cite{nguyen2022vindr}: This dataset aims to localize organs and nodules from 2D X-ray lung images. The classes include aortic enlargement, atelectasis, calcification, cardiomegaly, consolidation, ILD, infiltration, lung opacity, nodule/mass, pleural effusion, pleural thickening, pneumothorax, pulmonary fibrosis, and other lesions. The total of X-ray images has abnormal tissues is 4394 images. For images where labels of the same class obtained from different experts are overlapped, we pre-process by averaging overlapping bounding boxes with an intersection-over-union of $20\%$. The training, validation, and testing rate are presented in Table \ref{tab:downstream-dataset}. In this task, we build a model to detect all nodules available in testing images.
    
    \item \textit{JSRT} \cite{shiraishi2000development,van2006segmentation}: This dataset includes 2D X-ray images taken from lung organs. The annotations consist of the heart, left clavicle, right clavicle, left lung, and right lung. We construct models to segment all organs in testing images followed by baselines in \cite{xie2021unified}.  
\end{itemize}

\begin{table*}
\caption{Details of datasets for downstream tasks. For loss functions, \textit{Dice} indicates dice loss, \textit{CE} indicates cross-entropy loss, and \textit{L1} indicates L1-loss. 
}
\vspace{-0.15in}
\setlength\tabcolsep{3.0pt}
\label{tab:downstream-dataset}
\begin{center}
\scalebox{0.8}{
\begin{tabular}{l c c c c c c}
\toprule
Dataset & MMWHS-CT & MMWHS-MRI & BraTS2018 & LUNA2016-v2 & VinDr-CXR & JSRT\\
\midrule  
Task & 3D Segmentation & 3D Segmentation & 3D Segmentation & 3D Classification & 2D object detection & 2D Segmentation\\
Modality & 3D CT & 3D MRI & 3D MRI & 3D CT & 2D X-ray & 2D X-ray\\
Training samples& 13 & 13 & 133 & 377138 & 3075 & 114\\
Validation samples& 3 & 3 & 57 & 152362 & 440 & 10\\
Test samples& 4 & 4 & 95 & 225475 & 879 & 123\\
Mean Data Size& $224 \times 224 \times 265$ & $224 \times 224 \times 145$ & $224 \times 224 \times 155$ & $64 \times 64 \times 32$ & $512 \times 512$ & $224 \times 224$\\
Loss & Dice + CE & Dice + CE & Dice + CE & CE & L1 + CE & Dice + CE \\
\bottomrule
\end{tabular}}
\end{center}
\end{table*}
\vspace{-0.1in}
\subsection{\textbf{Other Settings}}
\paragraph{\textbf{2D $\&$ 3D Segmentation Tasks:}}
 We formulate 3D segmentation tasks in BraTS2018, MMWHS-CT/MRI as the 2D segmentation problem (JSRT data). To this end, we create two subsets to avoid imbalance issues. The first one includes 2D images whose labels contain a target object required to segment, and the other comprises background and remaining structures. We then sample  data from these two sets and train with a U-Net model. The backbone for this U-Net is the pre-trained ResNet-50 ($f_{\mathrm{intra}}$ in our model). We employ the SGD as the optimizer and  learning rate $0.1$ for segmentation-related tasks. All 2D SSL baselines are trained with $50$ epochs. Our method usually converges faster so we picked $15,\,25,\,50,\,50$ epochs for MMWHS-MRI, MMWHS-CT, BraTS2018, and JSST ($100\%$ case) respectively. The results are reported in average performance over five trial times using 3D Intersection over Union (IoU) for 3D settings and 2D Dice for 2D cases.
 
\paragraph{\textbf{3D Object classification (LUNA2016-v2)}}
Given the output feature maps of deformable attention, we build on top of the $f_{\textrm{intra}}$ and $f_{\textrm{inter}}$ two fully connected layers in the size of $640$ and $252$. The last layer returns two probability outputs for the binary classification problem. We use the binary cross entropy as the loss function and train with SGD using learning rate of $0.1$ for $100$ epochs. The Area Under the Curve (AUC) metric is used to evaluate performance. We report average performance for different batch sizes of $8, 16, 32$ each case with two trials.


\paragraph{\textbf{2D Object Detection (VinDr-CXR)}}
We choose the Faster R-CNN from MMDetection\footnote{\url{https://github.com/open-mmlab/mmdetection}} framework as a base model for the 2D object detection. This model is loaded pre-trained ResNet-50 derived from different 2D SSL methods. With 2D SSL baselines, the detector is trained with SGD as the optimizer with a learning rate $lr=0.001$ and converges after $25$ epochs. However, our pre-trained weights could not train with this learning rate (NAN loss), we thus selected $lr = 0.1$ for the first $30$ epochs and rescheduled $lr = 0.01$ for the last $10$ epochs to get stable checkpoints (normally converge after $35$ epochs). Other parameters are set as default by MMDetection. We use the mean average precision (mAP) of all classes with an IoU threshold fixed at 0.5 (mAP@0.5) for the evaluation as \cite{benvcevic2022self}. The results are computed in average of three trial times.


\section{Performance of Pre-trained Models when Varying Training Data}

\begin{figure}[H]
\centering
\includegraphics[width=0.5 \textwidth]{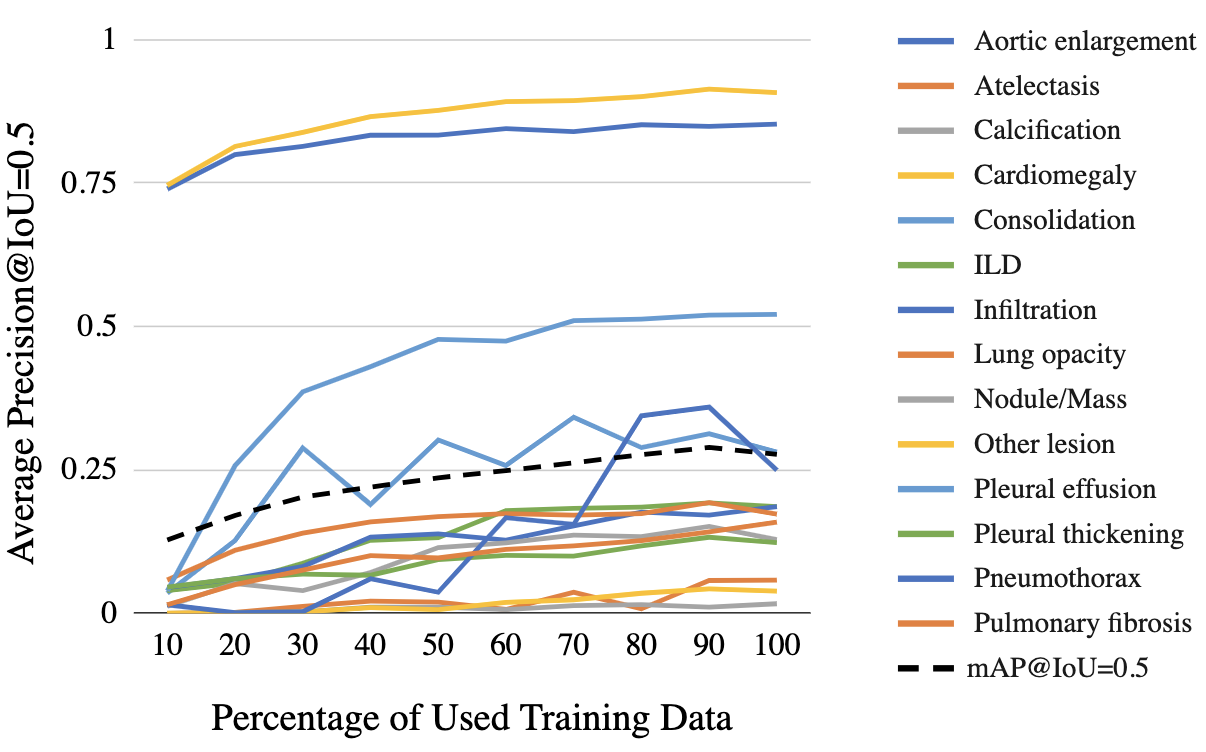}
\vspace{-0.2in}
\caption{\small{The mean average precision across different percentages of labeled data in abnormal Chest X-rays detection.}}
\label{fig:map-class-wise}
\end{figure}

In this experiment, we investigate
behaviors of our pre-trained model (extended from Deepcluster-v2) on downstream tasks when varying training data size. We conduct testings on the VinDr-CXR dataset by increasing training samples from 10$\%$ to 100$\%$ during the fine-tuning phase. Figure \ref{fig:map-class-wise} indicates class-wise average precision and the overall performance across classes (dashed black curve). As can be seen, there exists a trend of improving performance when more data is available; however, at the rate of 80$\%$, we already achieved a comparable accuracy of 90$\%$ or 100$\%$. This evidence suggests that the proposed SSL method can use fewer labeled data but still can achieve similar performance. We argue that this property is valuable, especially in medical applications where obtaining labeled instances is extremely expensive.

\section{More Visualization Results}
\begin{figure}[H]
\centering
\includegraphics[width=0.48\textwidth]{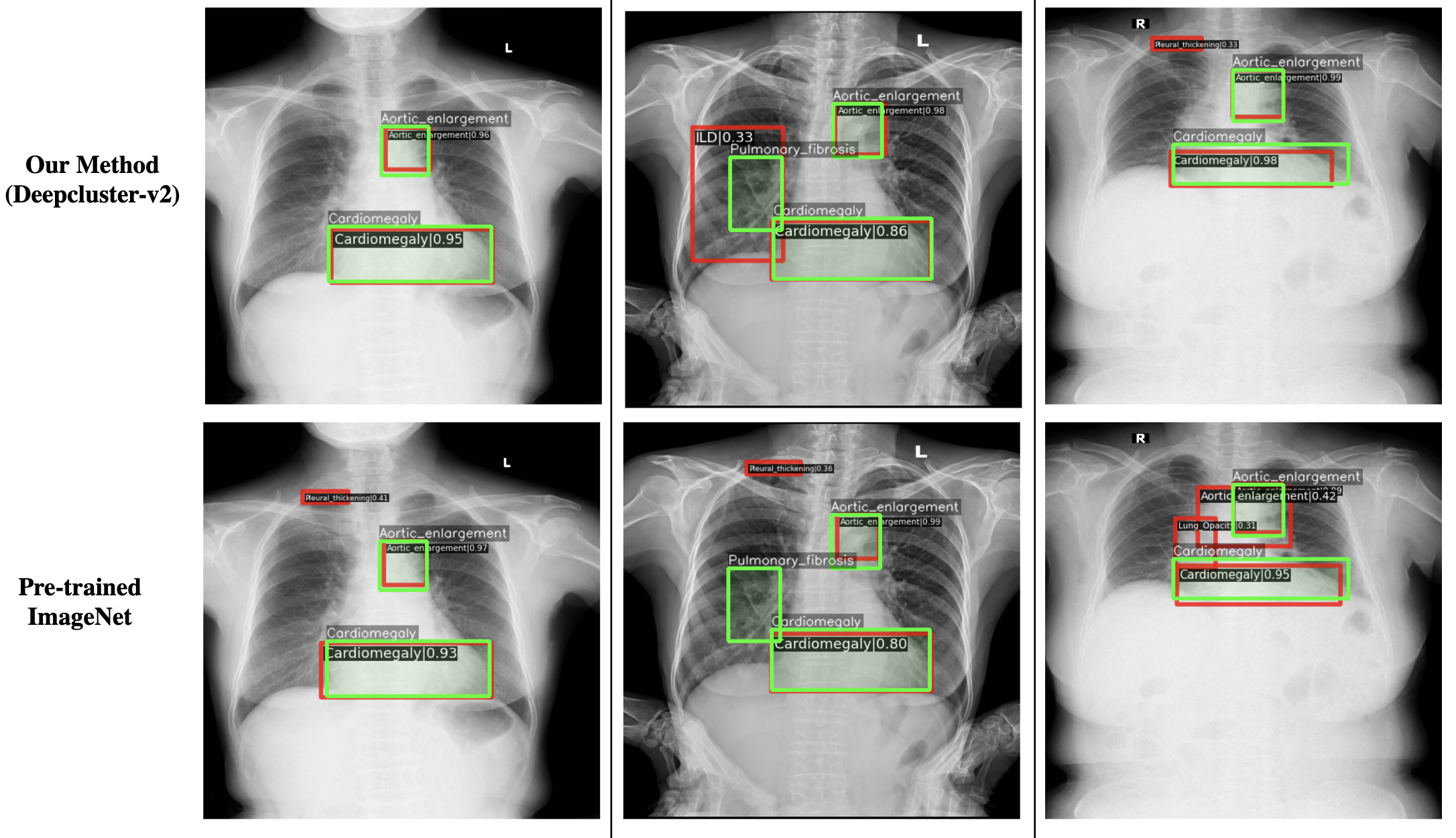}
\vspace{-0.2in}
\caption{\small{Qualitative results on detecting abnormal nodules in the VinDr-CXR dataset. Our method uses Deepcluster-v2, and the pre-trained Imagenet backbone is the first and second row. Green and red indicate ground truths and predictions, respectively.}}
\label{fig:vindr-cxr-more}
\end{figure}

Figure \ref{fig:vindr-cxr-more} provides qualitative results on detecting abnormal nodules 2D X-rays in the VinDr-CXR. We compare our extended version using Deepcluster-v2 with the pre-trained Imagenet. Across different testing images, it seems that the proposed method might help reduce false positive predictions. For instance, the baseline incorrectly signifies the pleural thickening structure in the first and second columns. In the third case, though our method also wrongly predicts pleural thickening, the errors are less than the baseline when there are wrong bounding boxes for either the lung opacity or aortic enlargement. 

\vspace{-0.15in}
\section{Current Limitations and Future Directions}
While the proposed method achieves state-of-the-art performance in several settings, the results on 2D object detection (VinDr-CXR) and 3D classification tasks (LUNA2016-v2) have not improved with large margins or still be smaller compared to other 3D SSL competitors. We argue that our learned features probably do not aid significantly these downstream tasks. We believe that further improvements might be gained by integrating downstream tasks’ properties in the pre-training algorithms as current insights in \cite{ericsson2021well,cole2022does}.

We also recommend conducting additional experiments on color images, such as skin attribute segmentation/classification \cite{nguyen2020visually,sun2021lesion}, diabetic retinopathy grading \cite{nguyen2021self,sun2021lesion}, or low-resolution images \cite{nguyen20173d,de2022deep}, to further validate the method's generalizations. Furthermore, it is critical to extending the framework with a similar mechanism for other SSL methods or learning under scenarios such as data imbalance or domain shift.

\end{document}